\documentclass[conference]{IEEEtran}
\IEEEoverridecommandlockouts
\usepackage{cite}
\usepackage{graphicx}
\usepackage{xcolor}
\usepackage{pifont}
\usepackage{colortbl}
\usepackage{zebra-goodies}
\usepackage{booktabs}
\usepackage{multirow}

\usepackage{algorithm}
\usepackage{algpseudocode}
\usepackage{amsmath}
\usepackage{amssymb}
\usepackage{url}

\usepackage{cleveref}
\crefname{figure}{Fig.}{Figs.}
\Crefname{figure}{Figure}{Figures}
\crefname{table}{Tab.}{Tabs.}
\Crefname{table}{Table}{Tables}

\usepackage{subcaption}
\usepackage{booktabs}

\def\BibTeX{{\rm B\kern-.05em{\sc i\kern-.025em b}\kern-.08em
    T\kern-.1667em\lower.7ex\hbox{E}\kern-.125emX}}
\begin{document}

\title{TraRA: \underline{Tra}jectory-level \underline{R}ecognition \underline{A}ggregation for Video Text Spotting in Urban Surveillance}


\author{
Duc Tri Tran$^{1,2}$,
Trung Thanh Nguyen$^{1,3}$,
Vijay John$^{4}$,
Phi Le Nguyen$^{2}$,
and Yasutomo Kawanishi$^{1,5}$
\thanks{Corresponding author. Email: kwns@fc.ritsumei.ac.jp}\\

$^{1}$RIKEN, Japan \hspace{10pt}
$^{2}$Hanoi University of Science and Technology, Vietnam\\
$^{3}$Nagoya University, Japan \hspace{10pt}
$^{4}$Lawrence Technological University, USA \hspace{10pt}
$^{5}$Ritsumeikan University, Japan
}

\maketitle

\begin{abstract}
Video Text Spotting (VTS) is essential for urban surveillance and intelligent transportation systems, enabling automated reading of street signs, vehicle markings, and scene text in video streams. However, reliable recognition remains challenging due to dynamic video factors common in surveillance scenarios, including motion blur, occlusion, and scale variation, which degrade frame-level recognition. Existing VTS methods typically perform recognition independently on each frame, leading to inconsistent and inaccurate results across sequences. To address these limitations, we propose TraRA (Trajectory-level Recognition Aggregation for VTS), a plug-and-play method that performs trajectory-level text recognition by leveraging temporal and multimodal consistency. TraRA integrates two key modules: (1) the Temporal Clustering and (2) the Vision-Language Aggregation. The former refines noisy trajectories by grouping temporally and visually coherent text instances, while the latter employs a Low-Rank Adaptation-enhanced Vision-Language model to fuse visual cues with linguistic context across frames. By aggregating information over entire text trajectories, TraRA achieves robust text recognition even under challenging surveillance conditions. Extensive experiments on four public benchmarks, including road and urban scene datasets (RoadText, BOVText, ArTVideo, and ICDAR15), demonstrate that TraRA consistently improves tracking and recognition performance over state-of-the-art VTS methods. The source code is available at \url{https://github.com/trid2912/TraRA}.
\end{abstract}

\begin{IEEEkeywords}
Temporal Clustering, Trajectory-level Recognition, Video Text Spotting, Vision-Language Model.
\end{IEEEkeywords}

\section{Introduction}
\label{sec:introduction}

The ability to robustly detect and recognize text in video streams is essential for urban surveillance systems~\cite{taki2023scene, long2021scene}, including autonomous driving, intelligent navigation, traffic monitoring, and general scene understanding. 
In these scenarios, automatically reading street signs, vehicle markings, and storefront text from surveillance footage enables richer situational awareness and supports downstream tasks such as vehicle identification and urban scene analysis. 
Video Text Spotting (VTS) addresses this need by extending static image text spotting into the temporal domain, involving localizing, tracking, and recognizing text that appears across a sequence of video frames. 
A common approach in modern VTS methods~\cite{wu2024end,he2024gomatching,he2025gomatching++} is to employ a state-of-the-art image text spotter~\cite{ye2023deepsolo} to detect text in each frame, followed by associating these frame-level detections across time using a robust tracker, following the tracking-by-detection paradigm.

\begin{figure}[t]
    \centering
    \begin{subfigure}[t]{0.9\linewidth}
        \centering
        \includegraphics[width=\linewidth]{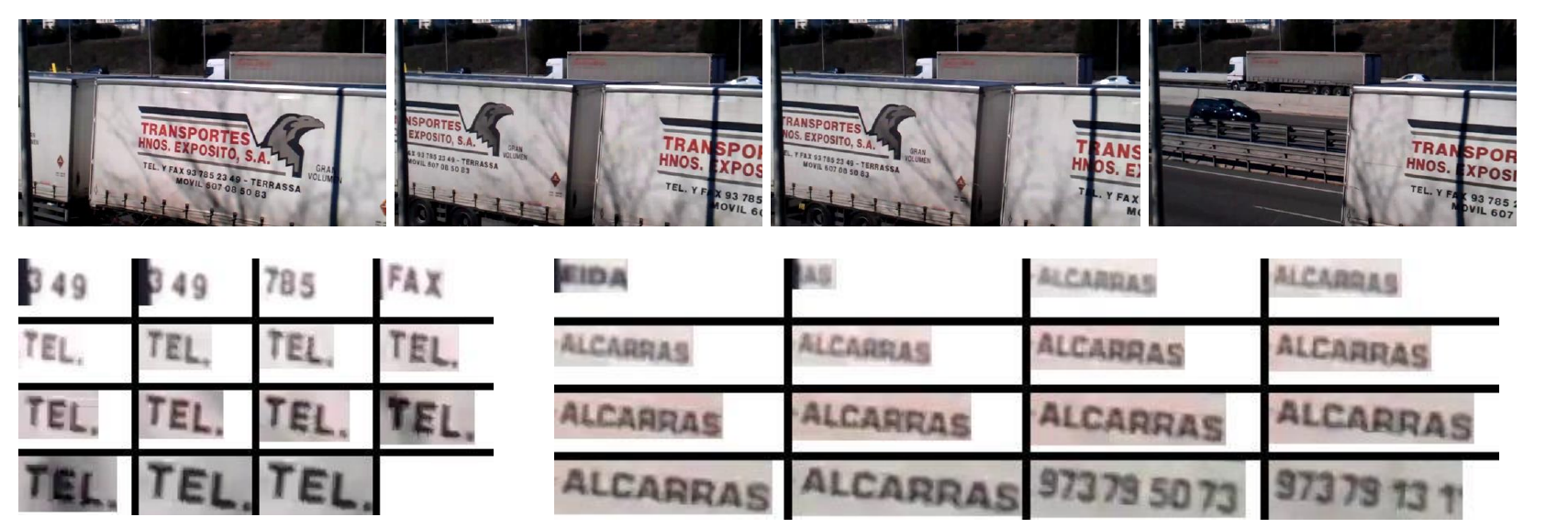}
        \caption{Identity switches from wrong cross-frame associations.}
        \label{fig:wrong_matching}
    \end{subfigure}
    \hfill
    \begin{subfigure}[t]{0.9\linewidth}
        \centering
        \includegraphics[width=\linewidth]{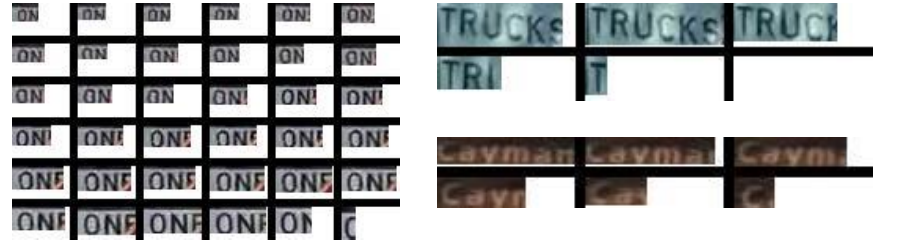}
        \caption{Incomplete trajectories due to fragmented tracking.}
        \label{fig:sequence_sample}
    \end{subfigure}

    \caption{Failure cases of the state-of-the-art VTS model GoMatching++~\cite{he2025gomatching++}.}
    \label{fig:vts_prob}
    \vspace{-1em}
\end{figure}

However, as shown in~\cref{fig:vts_prob}, surveillance videos are particularly susceptible to dynamic factors such as rapid camera or text movement and frequent occlusions, which often lead to unreliable text detection and recognition at the frame-level~\cite{Bautista2022SceneTR,Du2022SVTRST,ye2023deepsolo}.
Consequently, matching these degraded and inconsistent frame-wise predictions across the sequence can introduce substantial uncertainty and reduce overall accuracy of the final text recognition.
To mitigate these issues, many approaches enhance image-based text spotters by synthetically applying blur and other video-like degradations to static images~\cite{wu2021bilingual,wu2024dstext,wu2024end}.
Some methods also leverage synthetic video datasets~\cite{karatzas2015icdar,ye2025textssr} to improve robustness against the degraded text quality in real-world videos.
Yet, this strategy has an inherent limitation.
Excessive simulation of severe video conditions forces recognition models to learn from text that is often unrecognizable even to humans.

In this paper, we propose TraRA, a plug-and-play method for video text recognition that addresses the limitations of frame-based prediction in existing VTS methods.
TraRA integrates two key components: (1) Temporal Clustering (TC) and (2) Vision-Language Aggregation (VLA), which work together to improve trajectory consistency and word recognition accuracy.
The TC module refines the noisy trajectories produced by state-of-the-art VTS models by grouping visually and temporally consistent text instances into coherent trajectories.
Afterward, the VLA module, built upon a Vision-Language Model (VLM) enhanced with a lightweight Low-Rank Adaptation (LoRA)~\cite{hu2022lora,Chen_QAdapter}, processes the entire sequence of cropped text instances within each trajectory.
By combining fragmented visual cues with linguistic context across all frames, the VLM generates a unified and reliable word prediction that remains robust even under motion blur, occlusion, or low-resolution conditions typical of surveillance footage.
Through this joint temporal refinement and multimodal reasoning, TraRA improves trajectory-level word recognition and enhances the overall performance of VTS.
The main contributions of this work are summarized as follows:
\begin{itemize}
    \item We propose TraRA, a plug-and-play method for VTS that enhances trajectory-level text recognition by jointly leveraging temporal consistency through the Temporal Clustering (TC) module and multimodal reasoning via the Vision-Language Aggregation (VLA) module.
    \item The TC module refines noisy trajectories by grouping temporally and visually consistent text instances, while the VLA module, based on LoRA-enhanced VLMs, integrates visual and linguistic cues for robust word recognition.
    \item Extensive experiments on four VTS benchmarks, including road and urban scene datasets, demonstrate that TraRA consistently improves tracking accuracy and recognition performance over state-of-the-art VTS methods.
\end{itemize}
%

\section{Related Work}
\label{sec:related_work}

\noindent \textbf{Scene Text Recognition (STR).}  
Research on STR has evolved from CNN-RNN hybrid frameworks~\cite{Shi2015AnET} to Transformer-based architectures~\cite{Bautista2022SceneTR,Du2022SVTRST,Fang2021ReadLH}, with a clear shift toward attention-driven models that capture global contextual features and complex character relationships. Notably, ABINet~\cite{Fang2021ReadLH} introduced an iterative refinement mechanism between vision and language modules, inspiring subsequent methods such as SVTR~\cite{Du2022SVTRST} and PARSeq~\cite{Bautista2022SceneTR}, which unify feature extraction and sequence modeling for improved accuracy. Contemporary research further extends these advances through lightweight architectures~\cite{Cheng2024SVIPTRFA}, diffusion-based models~\cite{Fujitake2023DiffusionSTRDM}, and synthetic data generation~\cite{ye2025textssr}. Despite these advancements, most STR methods remain fundamentally frame-level and do not exploit temporal cues across frames, treating video sequences as independent snapshots without leveraging inter-frame continuity.

\vspace{5pt}
\noindent \textbf{Video Text Spotting (VTS).} 
The field of VTS, which unifies text detection and recognition in videos, has advanced considerably through image-based methods~\cite{liao2020mask,huang2022swintextspotter,ye2023deepsolo}, but faces additional challenges such as flicker and temporal inconsistency. 
Early methods incorporated text tracking to move beyond frame-by-frame processing. Cheng et al.~\cite{Cheng2020FREEAF} introduced FREE, which leveraged temporal cues to stabilize recognition by selecting the highest-quality text instances across sequences. 
With the rise of Transformer architectures, Wu et al.~\cite{wu2024end} introduced TransDeTR, which reframed VTS as long-range sequence learning, jointly integrating detection, tracking, and recognition in an end-to-end framework. 
More recent work, such as GoMatching~\cite{he2024gomatching} and GoMatching++~\cite{he2025gomatching++}, augments the state-of-the-art image spotter DeepSolo~\cite{ye2023deepsolo} with a customized tracking module for end-to-end video text spotting. 
Despite these advances, most existing methods still rely on frame-level results, where a single misread frame can corrupt the entire text trajectory.

\vspace{5pt}
\noindent \textbf{Vision-Language Models (VLMs).}  
Recent advances in VLMs have enabled unified reasoning across visual and textual modalities, leading to a new generation of OCR frameworks.
Traditional pipelines that separate detection and recognition are increasingly being replaced by multimodal architectures capable of contextual understanding. 
Zhao et al.~\cite{Zhao2023CLIP4STRAS} proposed CLIP4STR, which adapts CLIP~\cite{Radford2021LearningTV} for scene text recognition, while Nougat et al.~\cite{Blecher2023NougatNO} introduced an end-to-end framework for academic document OCR. 
Inspired by these developments, we leverage Ovis2.5~\cite{lu2025ovis2} for VTS, given its strong vision-language alignment and robust temporal reasoning capabilities as evidenced by its performance on the OpenCompass leaderboard~\cite{duan2024vlmevalkit}. 
This integration enables the proposed method to capture semantic coherence across frames, thereby addressing the limitations of traditional frame-level recognition strategies.

\section{Proposed TraRA Method}
\label{sec:proposed_method}

\begin{figure*}[t]
    \centering
    \begin{subfigure}[t]{0.59\textwidth}
        \centering
        \includegraphics[width=\linewidth]{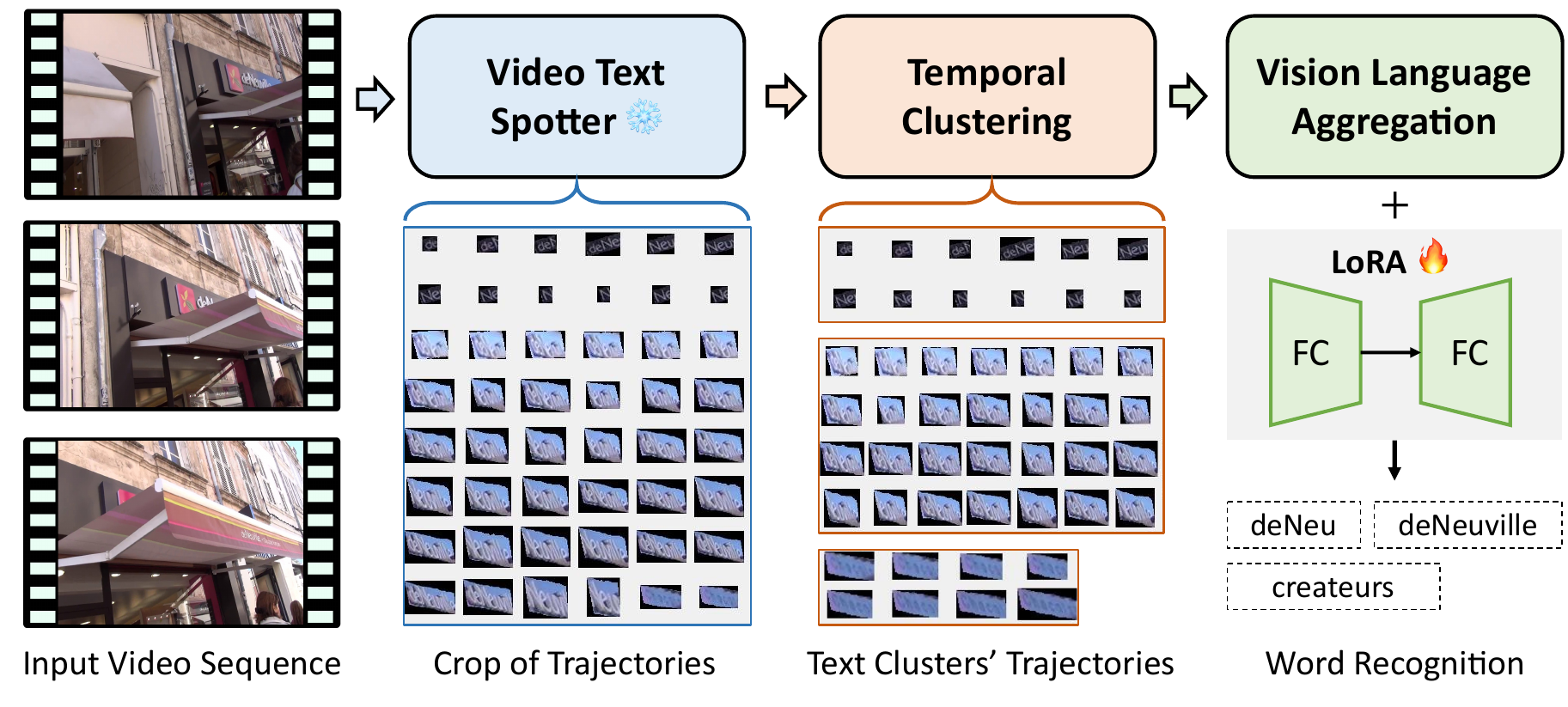}
        \caption{Overview of the proposed TraRA. The video text spotter extracts trajectory-level text crops, which are refined by Temporal Clustering and Vision-Language Aggregation for word recognition.}
        \label{fig:pipeline}
    \end{subfigure}
    \hfill
    \begin{subfigure}[t]{0.38\textwidth}
        \centering
        \includegraphics[width=\linewidth]{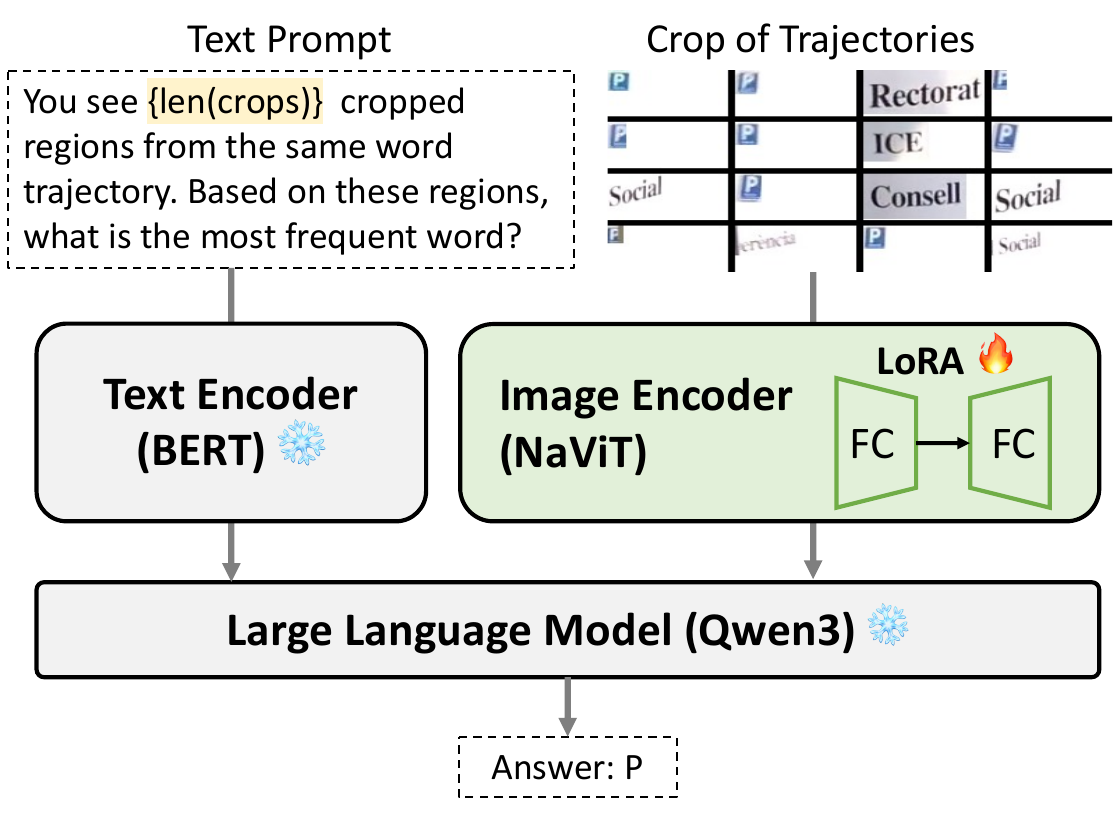}
        \caption{
        Parameter-efficient fine-tuning strategy. 
        The entire model is frozen except for the last block of the Image Encoder.
        }
        \label{fig:vlm}
    \end{subfigure}
    \caption{Overview of the proposed TraRA and its Vision-Language Model fine-tuning strategy for video text spotting.}
    \label{fig:architecture}
\end{figure*}

\Cref{fig:architecture} presents an overview of the proposed TraRA for video text spotting.
Given an input video sequence, a pre-trained Video Text Spotter (VTS) detects and tracks text regions across frames to produce trajectory-level text crops. However, these trajectories are often prone to identity switches, as illustrated in~\cref{fig:wrong_matching}.
To refine these results, we introduce the Temporal Clustering (TC) module, which groups visually and temporally consistent instances into coherent trajectories.
As shown in~\cref{fig:sequence_sample}, another major challenge arises from incomplete or obscured frames, where traditional majority voting fails because most frames do not contain the full word.
To address this, we propose the Vision-Language Aggregation (VLA) module, enhanced with lightweight Low-Rank Adaptation (LoRA)~\cite{hu2022lora}.
This module performs word recognition across the entire sequence by integrating fragmented visual cues with linguistic context, producing accurate predictions even when individual frames are unreliable.
By jointly leveraging temporal consistency and multimodal reasoning, TraRA improves trajectory-level word recognition and corrects erroneous predictions from the VTS model.

\vspace{5pt}
\noindent \textbf{Problem Formulation.} 
Let $I = [I_1, I_2, \dots, I_T]$ denote a video sequence of $T$ frames. 
A VTS model $f_{\text{VTS}}$ produces a set of text trajectories 
$S = \{t_1, t_2, \dots, t_N\}$ with $S = f_{\text{VTS}}(I)$ 
and corresponding recognition results 
$R = \{r_1, r_2, \dots, r_N\}$. 
Each trajectory $t_i = [x_1, x_2, \dots, x_{M_i}]$ represents a sequence of $M_i$ detected text instances across frames.
The proposed TraRA refine these trajectories through two modules. 
First, Temporal Clustering $f_{\text{TC}}$ segments each trajectory into temporally consistent sub-trajectories 
$Q_i = \{t_i^1, t_i^2, \dots\}$, where $Q_i = f_{\text{TC}}(t_i)$. 
Second, Vision-Language Aggregation $f_{\text{VLA}}$ generates the recognition output for each sub-trajectory as 
$r_i^k = f_{\text{VLA}}(t_i^k)$. 
The updated trajectory and recognition sets are as follows:
\begin{equation}
    S_{\text{update}} = \{t_1^1, t_1^2, \dots, t_N^{K_N}\}, 
    \quad 
    R_{\text{update}} = \{r_1^1, r_1^2, \dots, r_N^{K_N}\}. \nonumber
\end{equation}
This ensures that each trajectory is temporally coherent and semantically refined.

\subsection{Temporal Clustering}
Existing VTS methods often suffer from trajectory matching errors when detection boxes from different text instances share similar embedding representations, leading to incorrect cross-frame associations.
To address this, we propose Temporal Clustering (TC), which refines noisy tracklets through two stages: (1) Feature Extraction \& Adaptive Threshold Identification and (2) Time-based Clustering, as shown in Algorithm~\ref{alg:clean_tracks}.
In the first stage, we extract visual features from masked text crops, where VTS prediction masks isolate text pixels and background pixels are set to zero. 
These features are used to compute an adaptive distance threshold $\tau$ based on the mean feature variation within the tracklet, dynamically adjusting to motion, blur, or appearance changes.
In the second stage, an online clustering process sequentially assigns each detection to the nearest existing cluster or initializes a new one when the feature distance exceeds $\tau$, thereby splitting inconsistent trajectories while preserving temporal coherence.

\begin{algorithm}[t]
\caption{Temporal Clustering Algorithm} 
\label{alg:clean_tracks}
\small
\begin{algorithmic}[1]
    \For{each original tracklet}
        \State Let the tracklet contain $n$ detections
        \State Extract masked crops for all detections in the tracklet
        \State \textbf{// Feature Extraction \& Adaptive Threshold Identification}
        \State Extract features for each crop:
        \State \hspace{20pt} $\mathbf{X}_i = [\, f_{\text{HOG}} ;\, f_{\text{SIFT}} ;\, A_i \,]$, $i = \{1,\dots,n\}$
        \State Compute pairwise feature distances
        \State \hspace{20pt} $d_i = \|\mathbf{X}_{i+1} - \mathbf{X}_i\|_2$, $i = \{1,\dots,n{-}1\}$, 
        \State \hspace{20pt} to obtain $D = [d_1, d_2, \dots, d_{n-1}]$
        \State Compute adaptive threshold $\tau = \max(\alpha \cdot \mathrm{mean}(D),\, \beta)$ 
        \State \Comment{$\alpha$, $\beta$ are tunable}
        \State \textbf{// Time-based Clustering}
        \State Initialize an empty list of active clusters $\mathcal{C}$
        \For{each feature vector $\mathbf{X}_i$ in temporal order}
            \State Find cluster $c \in \mathcal{C}$ minimizing: $d(\mathbf{X}_i, c) = \|\mathbf{X}_i - \mathbf{x}_{\mathrm{last}}^{(c)}\|_2$
            \If{$d(\mathbf{X}_i, c) < \tau$}
                \State Append $\mathbf{X}_i$ to cluster $c$
            \Else
                \State Initialize a new cluster with $\mathbf{X}_i$
            \EndIf
        \EndFor
        \State Update trajectory set $\mathcal{C}$ and assign new global IDs
    \EndFor
\end{algorithmic}
\end{algorithm}

\subsubsection{Feature Extraction \& Adaptive Threshold Identification}
Existing query-based matching methods often produce similar representations for different text fragments sharing semantic traits, causing matching ambiguities across trajectories. 
They also tend to ignore the physical scale of crops, leading to trajectory drifting. 
To address this, we design a discriminative feature representation for masked text crops.
Each detection is represented by $\mathbf{X}_i = [\, f_{\text{HOG}} ;\, f_{\text{SIFT}} ;\, A_i \,]$, integrating three complementary components:
\begin{itemize}
    \item {Histogram of Oriented Gradients (HOG)}: For a grayscale crop $I(x, y)$, we compute gradients $G_x = \frac{\partial I}{\partial x}$, $G_y = \frac{\partial I}{\partial y}$, with magnitude $M(x, y) = \sqrt{G_x^2 + G_y^2}$ and orientation $\theta(x, y) = \tan^{-1}\!\left(\frac{G_y}{G_x}\right)$. Orientation histograms weighted by $M$ are accumulated per cell and block-normalized to produce $f_{\text{HOG}}$, encoding structural and textural variations robust to photometric changes.

    \item Scale-Invariant Feature Transform (SIFT): Each crop is resized to $64 \times 64$, and keypoints $\{p_k\}$ are densely sampled on a regular grid. For each keypoint, a 128-dimensional descriptor $f^k_{\text{SIFT}} \in \mathbb{R}^{128}$ encodes gradient orientations within a $16 \times 16$ region using $4 \times 4$ subregion histograms with 8 bins. Descriptors are converted to RootSIFT via $\ell_1$-normalization and element-wise square rooting, then averaged and $\ell_2$-normalized to form $f_{\text{SIFT}}$.

     \item {Area}: $A_i = w_i \times h_i$ captures the bounding-box area, providing scale context to identify abrupt size variations indicating mismatched associations.

\end{itemize}
After computing $\mathbf{X}_i$ for all detections, we estimate an adaptive threshold $\tau$ from consecutive pairwise distances $d_i = \|\mathbf{X}_{i+1} - \mathbf{X}_i\|_2$, yielding $D = [d_1, d_2, \dots, d_{n-1}]$. The threshold is defined as:
\begin{equation}
    \tau = \max(\alpha \cdot \mathrm{mean}(D), \, \beta),
\end{equation}
where $\alpha$ and $\beta$ control sensitivity to intra-tracklet variation. 
Tracklets are split whenever consecutive feature distance exceeds $\tau$, isolating corrupted or mismatched segments.

\subsubsection{Time-based Clustering} 
Traditional clustering algorithms like k-means and DBSCAN are ill-suited for sequential video data because they lack a temporal dimension. 
To address these issues, we propose a Time-based Clustering that adapts the core principles of density-based clustering while processing data strictly in temporal order, ensuring temporal consistency in trajectory formation.
The proposed method operates sequentially, leveraging the inherent temporal structure of video data as follows:
\begin{itemize}
    \item {Online Processing:} Feature vectors $\mathbf{X}_i$ from each tracklet are processed in temporal order, maintaining a dynamic list of active clusters $\mathcal{C}$.
    \item {Adaptive Threshold Identification:} Prior to clustering, an adaptive distance threshold $\tau$ is computed, providing a trajectory-specific boundary for feature consistency.
    \item {Cluster Assignment:} 
    For each feature vector $\mathbf{X}_i$, we identify the nearest active cluster $C \in \mathcal{C}$ by computing the $\ell_2$ distance to its most recent entry:
    \begin{equation}
        d(\mathbf{X}_i, C) = \|\mathbf{X}_i - \mathbf{x}_{\mathrm{last}}^{(C)}\|_2,
    \end{equation}
    where $\mathbf{x}_{\mathrm{last}}^{(C)}$ represents the latest feature vector assigned to cluster $C$. If $d(\mathbf{X}_i, C) < \tau$, the vector $\mathbf{X}_i$ is appended to $C$ and becomes the new reference point for that cluster; otherwise, a new cluster is initialized with $\mathbf{X}_i$ as its first member.
\end{itemize}
The proposed Time-based Clustering, guided by the adaptive threshold $\tau$, ensures that the resulting clusters $\mathcal{C}$ remain temporally contiguous and feature-consistent.

\subsection{Vision-Language Aggregation}
We employ Ovis2.5-9B~\cite{lu2025ovis2} as the VLM
\footnote{Ovis2.5-9B~\cite{lu2025ovis2} is a state-of-the-art VLM at the time of this work.} 
backbone for TraRA. 
The model consists of two components: a SigLIP2 image encoder\cite{tschannen2025siglip} with NaViT-style variable-resolution processing\cite{NEURIPS2023_06ea400b} that extracts visual tokens from text crops, and a Qwen3-8B large language model~\cite{qwen3} that receives the visual tokens together with a tokenized text prompt (e.g., ``What is the full word of this image?'') and produces the final word prediction.

\subsubsection{Fine-tuning} 
We adopt a parameter-efficient fine-tuning strategy based on LoRA~\cite{hu2022lora} to adapt the VLM to video text data while minimizing computational cost, as illustrated in \cref{fig:vlm}. All model parameters are frozen except for LoRA adapters inserted into the last Transformer block of the SigLIP2 image encoder (layer 26 of 27). We use rank $r{=}32$, scaling factor $\alpha{=}64$, and dropout $p{=}0.05$, yielding approximately 0.64M trainable parameters out of 9B total ($<$0.01\% of the model).
During fine-tuning, the input consists of a mixture of correct text crops and distractor crops sampled from unrelated trajectories. 
By prompting the model to identify the most frequent word across this heterogeneous set, the VLM learns to discriminate and aggregate text instances, thereby mitigating inter-trajectory ambiguity.

\subsubsection{Inference} 
After fine-tuning, the VLM serves as a trajectory-level word aggregator. 
For each clustered trajectory, all associated crops are sequentially encoded and passed through the fine-tuned VLM. 
The decoder aggregates these embeddings to produce the final word prediction, integrating temporal and semantic consistency across the entire trajectory.

\section{Evaluation}
\label{sec:experiments}
\subsection{Experimental Conditions}
\noindent \textbf{Datasets.}
We evaluate TraRA on four public VTS benchmarks: ArTVideo~\cite{he2024gomatching} for arbitrarily shaped text, ICDAR15~\cite{karatzas2015icdar} for incidental scene text, RoadText~\cite{reddy2020roadtext} for road signs and vehicle markings, and BOVText~\cite{wu2021bilingual} for large-scale bilingual videos.

\vspace{5pt}
\noindent \textbf{Evaluation Metrics.}
Following the protocols established in \cite{karatzas2015icdar, wu2021bilingual}, we adopt standard end-to-end Video Text Spotting (VTS) metrics to evaluate spatial-temporal consistency: {Multiple Object Tracking Accuracy (MOTA)}, {Multiple Object Tracking Precision (MOTP)} \cite{bernardin2008evaluating}, and the {Identification F1 Score (ID$_{\text{F1}}$)} \cite{ristani2016performance}. Within this framework, a prediction is considered a true positive only if both the trajectory association and the underlying text recognition are concurrently accurate.

While tracking metrics assess identity maintenance over time, recognition quality is a key driver of temporal stability. 
To isolate its contribution, we include Word Accuracy (WA) and Normalized Edit Distance (NED), following recent VTS frameworks \cite{wu2021bilingual} that treat recognition precision as a proxy for temporal consistency.

\vspace{5pt}
\noindent \textbf{Models \& Hyperparameters.}
We implement the proposed TraRA as a plug-and-play module within two state-of-the-art VTS methods, GoMatching++~\cite{he2025gomatching++} and TransDETR~\cite{wu2024end}.
For GoMatching++~\cite{he2025gomatching++}, we adopt the publicly released pretrained weights for ArTVideo, ICDAR15, and BOVText, and fine-tune the model on RoadText to obtain the reported results.
For TransDETR~\cite{wu2024end}, we use the pretrained model and perform inference on each dataset without additional fine-tuning. 
All baseline hyperparameters follow their original configurations.
For the Temporal Clustering module, we fix $\alpha$ at 1 and tune the hyperparameter $\beta$ for each dataset.

\subsection{Experimental Results}
\begin{table}[t]
\centering
\caption{
Quantitative comparison on multiple benchmarks using different VTS. 
Bold values indicate the best performance for each benchmark.
}
\label{tab:main_result_merged}
\setlength\tabcolsep{3pt}
\resizebox{\linewidth}{!}
{
\begin{tabular}{cccccccc}
\toprule
{VTS} & {Dataset} & {TraRA} & {WA$\uparrow$} & {NED$\uparrow$} & {MOTA$\uparrow$} & {MOTP$\uparrow$} & {ID$_\text{F1}$$\uparrow$} \\
\midrule

\multirow{8}{*}{\rotatebox[origin=c]{90}{{GoMatching++}}} 
& \multirow{2}{*}{ArTVideo} 
&  & 90.52 & \textbf{97.28} & 69.39 & \textbf{82.87} & 78.10 \\ 
& & \ding{51} & \textbf{92.19} & 95.50 & \textbf{73.93} & 82.62 & \textbf{80.40} \\ 
\cmidrule(l){2-8}

& \multirow{2}{*}{ICDAR15} 
&  & -- & -- & 72.20 & \textbf{78.52} & 80.11 \\ 
& & \ding{51} & -- & -- & \textbf{73.08} & 78.44 & \textbf{80.61} \\ 
\cmidrule(l){2-8}

& \multirow{2}{*}{RoadText} 
&  & 79.35 & \textbf{94.18} & 3.91 & 35.20 & 36.20 \\ 
& & \ding{51} & \textbf{81.89} & 93.28 & \textbf{4.43} & \textbf{35.50} & \textbf{36.30} \\ 
\cmidrule(l){2-8}

& \multirow{2}{*}{BOVText} 
&  & 74.67 & 82.76 & 52.90 & \textbf{87.20} & 62.80 \\ 
& & \ding{51} & \textbf{79.32} & \textbf{93.20} & \textbf{70.90} & 86.80 & \textbf{70.50} \\ 
\midrule

\multirow{4}{*}{\rotatebox[origin=c]{90}{TransDETR}} 
& \multirow{2}{*}{RoadText} 
&  & 25.81 & 61.50 & -11.20 & 25.20 & 28.70 \\ 
& & \ding{51} & \textbf{76.30} & \textbf{92.14} & \textbf{2.70} & \textbf{35.70} & \textbf{37.00} \\ 
\cmidrule(l){2-8}

& \multirow{2}{*}{BOVText} 
&  & 17.09 & 38.10 & -72.90 & 71.40 & 10.20 \\ 
& & \ding{51} & \textbf{64.94} & \textbf{81.70} & \textbf{-34.70} & \textbf{72.50} & \textbf{28.70} \\ 
\bottomrule
\end{tabular}
}
\end{table}

\noindent \textbf{Quantitative Results.}
As shown in~\cref{tab:main_result_merged}, integrating TraRA consistently improves performance across all benchmarks and VTS backbones. 
We report both recognition metrics (WA and NED), alongside tracking metrics (MOTA, MOTP, and ID$_\text{F1}$). 
When combined with GoMatching++~\cite{he2025gomatching++}, TraRA improves recognition accuracy WA on all evaluated datasets: +1.67 on ArTVideo, +2.54 on RoadText, and +4.65 on BOVText. 
The largest gains appear on BOVText, where NED also increases by +10.44, indicating substantially more accurate character-level predictions. 
Tracking metrics follow the same trend, with MOTA improving by up to +18.0 on BOVText and ID$_\text{F1}$ by +7.7, while MOTP remains stable across all datasets, confirming no degradation in localization precision. 
For TransDETR~\cite{wu2024end} under zero-shot settings, TraRA yields dramatically larger improvements. 
WA increases by +50.49 on RoadText and +47.85 on BOVText, with corresponding NED gains of +30.64 and +43.60. 
Tracking metrics improve accordingly, with MOTA gains of +13.9 and +38.2, respectively. 
These substantial improvements on a weaker baseline demonstrate that TraRA's trajectory-level aggregation is especially effective at compensating for unreliable frame-level recognition.





\begin{table}[t]
\centering
\caption{Ablation study of the proposed TraRA components using GoMatching++~\cite{he2025gomatching++} as the VTS. 
Bold values indicate the best performance.
}
\label{tab:ablation}
\setlength\tabcolsep{4pt} 
\resizebox{0.8\linewidth}{!}
{
\begin{tabular}{lcccccc}
\toprule
{Dataset} & {TC} & {VLA} & {MOTA$\uparrow$} & {MOTP$\uparrow$} & {ID$_\text{F1}$$\uparrow$} \\
\midrule
\multirow{4}{*}{ArTVideo}
& \ding{51} & \ding{51} & \textbf{73.93} & 82.62 & \textbf{80.40} \\ 
& \ding{51} &  & 69.38 & \textbf{82.87} & 78.08 \\ 
&  & \ding{51} & 72.90 & 82.58 & 80.36 \\ 
&  &  & 69.39 & \textbf{82.87} & 78.10 \\ 
\midrule
\multirow{4}{*}{ICDAR15}
& \ding{51} & \ding{51} & \textbf{73.08} & 78.44 & \textbf{80.61} \\ 
& \ding{51} &  & 72.21 & \textbf{78.52} & 80.11 \\ 
&  & \ding{51} & 73.02 & 78.44 & 80.60 \\ 
&  &  & 72.20 & \textbf{78.52} & 80.11 \\ 
\bottomrule
\end{tabular}
}
\end{table}

\begin{table}[t]
\centering
\caption{
Ablation study on the effect of different $\beta$ values in the TC module with $\alpha = 1$. 
Results are reported for TC-only and the full TraRA model on the ArTVideo~\cite{he2024gomatching} dataset using GoMatching++~\cite{he2025gomatching++} as the VTS. 
Bold values indicate the best performance.
}
\label{tab:ablation_cluster}
\setlength\tabcolsep{4pt}
\resizebox{\linewidth}{!}
{
\begin{tabular}{lccccccc}
\toprule
\multirow{2}{*}{$\alpha$} & \multirow{2}{*}{$\beta$} & \multicolumn{3}{c}{{TC-only}} & \multicolumn{3}{c}{{TraRA}} \\
\cmidrule(lr){3-5} \cmidrule(lr){6-8}
 &  & {MOTA$\uparrow$} & {MOTP$\uparrow$} & {ID$_\text{F1}$$\uparrow$} 
 & {MOTA$\uparrow$} & {MOTP$\uparrow$} & {ID$_\text{F1}$$\uparrow$} \\ 
\midrule
1.0 & 0.2 & 57.83 & 82.87 & 53.15 & 61.25 & \textbf{82.70} & 54.92  \\
1.0 & 0.4 & 68.98 & 82.87 & 77.69 & 73.00 & \textbf{82.70} & 79.75  \\
1.0 & 0.6 & 69.38 & 82.87 & 78.08 & \textbf{73.93} & 82.62 & \textbf{80.40}  \\
1.0 & 0.8 & \textbf{69.39} & 82.87 & \textbf{78.10} & 72.83 & 82.59 & 80.33  \\
1.0 & 1.0 & \textbf{69.39} & 82.87 & \textbf{78.10} & 72.90 & 82.58 & 80.36  \\
\bottomrule
\end{tabular}
}
\end{table}

\vspace{5pt}
\noindent \textbf{Ablation Studies.}
We analyze the contributions of TraRA's two key modules: the Temporal Clustering (TC) and the Vision-Language Aggregation (VLA). 
As shown in~\cref{tab:ablation}, VLA is the primary driver of improvement, it yields +3.51 of MOTA and +2.26 of ID$_\text{F1}$ on ArTVideo, with similar trends on ICDAR15 (+0.82 of MOTA and +0.49 of ID$_\text{F1}$). 
TC alone yields minimal change in isolation, but provides meaningful gains when combined with VLA, further boosting MOTA by +1.03 and ID$_\text{F1}$ by +0.04 on ArTVideo, and MOTA by +0.06 and ID$_\text{F1}$ by +0.01 on ICDAR15. 
This suggests that TC's trajectory refinement primarily benefits VLA by providing cleaner input sequences for aggregation.

We further evaluate the sensitivity of the TC module to the hyperparameter $\beta$, with $\alpha$ fixed at 1.0. 
As shown in~\cref{tab:ablation_cluster}, small $\beta$ values (e.g., 0.2) severely over-segment trajectories, degrading MOTA and ID$_\text{F1}$. 
Performance plateaus at $\beta=0.6$, where the proposed TraRA best balances temporal continuity and feature discrimination.

\begin{figure}[t]
    \centering
    \begin{subfigure}[t]{0.99\linewidth}
        \centering
        \includegraphics[width=\linewidth]{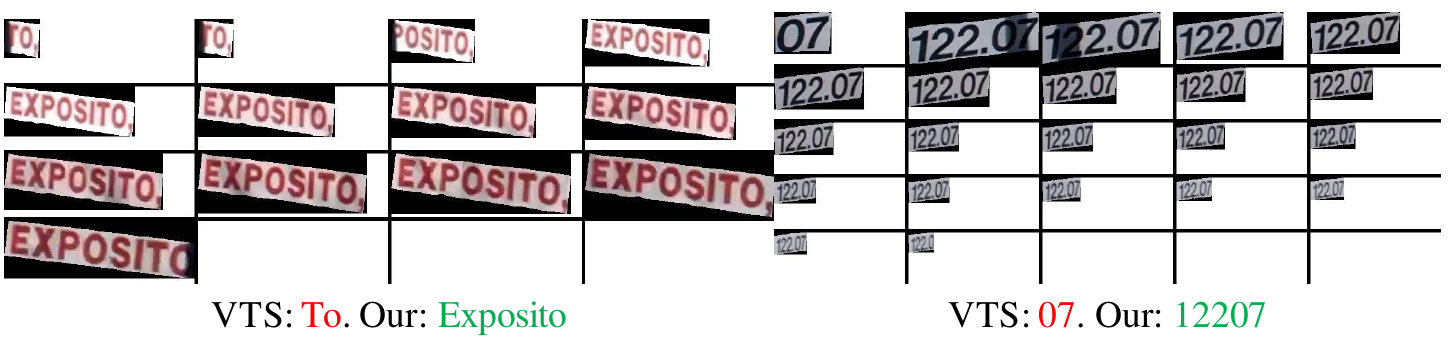}
        \caption{Example of successful recognition.}
        \label{fig:goodcase}
    \end{subfigure}
    \vspace{2mm}
    \begin{subfigure}[t]{0.85\linewidth}
        \centering
        \includegraphics[width=\linewidth]{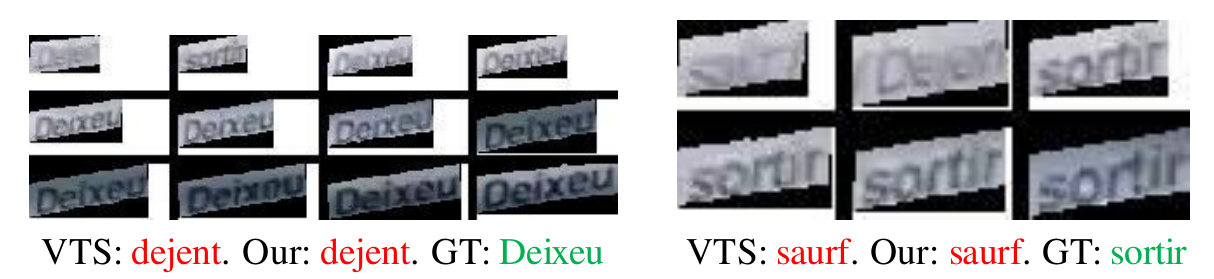}
        \caption{Example of recognition failure.}
        \label{fig:badcase}
    \end{subfigure}
    \caption{
        Qualitative comparison between the proposed TraRA (Our) and the state-of-the-art VTS (GoMatching++~\cite{he2025gomatching++}).
    }
    \label{fig:good_bad_cases}
\end{figure}

\vspace{5pt}
\noindent \textbf{Qualitative Results.}
\cref{fig:good_bad_cases} provides qualitative examples of the proposed TraRA in comparison with GoMatching++~\cite{he2025gomatching++}.
In \cref{fig:goodcase}, TraRA accurately reconstructs complete text sequences such as ``Exposito'' and ``12207'', where the GoMatching++~\cite{he2025gomatching++} produces incomplete predictions. 
This improvement stems from aggregating visual and linguistic cues across all frames, allowing recovery of missing characters under motion blur or occlusion. 
However, as shown in \cref{fig:badcase}, recognition errors may still occur in severely blurred or low-quality regions, leading to visually similar but incorrect predictions. 
Overall, TraRA consistently produces more complete and temporally consistent word recognition results than the GoMatching++~\cite{he2025gomatching++}.

\vspace{5pt}
\noindent \textbf{Limitation.}
Despite the consistent improvements achieved by TraRA across multiple benchmarks and VTS backbones, several limitations remain.
First, the TC module processes detections in an online fashion without long-term temporal memory or look-ahead, which ensures low latency but may struggle to recover trajectories during prolonged occlusion or complete text disappearance.
Second, while the VLA module effectively fuses visual and linguistic cues for robust recognition, integrating a VLM increases computational overhead compared to simple majority voting, making real-time inference challenging on standard hardware.
These limitations motivate future work on lightweight VLMs and efficient temporal modeling to maintain TraRA's recognition advantages while improving scalability.

\section{Conclusion}
\label{sec:conclusion}

In this paper, we proposed TraRA, a plug-and-play method for Video Text Spotting (VTS) that enhances trajectory-level text recognition through Temporal Clustering (TC) and Vision-Language Aggregation (VLA).
The TC module refines noisy and fragmented trajectories by grouping temporally and visually consistent detections.
The VLA module, powered by a LoRA-enhanced VLM, fuses visual cues and linguistic context across frames to achieve robust word recognition.
Experiments on four public benchmarks show that TraRA consistently improves both tracking stability and recognition accuracy over state-of-the-art VTS methods.
In future work, we plan to explore lightweight VLM architectures and end-to-end training strategies to further enhance temporal reasoning and enable real-time performance.

\section*{Acknowledgment}
This work was partly supported by the Japan Society for the Promotion of Science (JSPS) KAKENHI Numbers JP24H00733.

\bibliographystyle{IEEEtran}
\bibliography{references}

\end{document}